\documentclass[10pt,twocolumn,letterpaper]{article}

 \usepackage[pagenumbers]{cvpr} 

\definecolor{cvprblue}{rgb}{0.21,0.49,0.74}
\usepackage[pagebackref,breaklinks,colorlinks,allcolors=cvprblue]{hyperref}
\usepackage{multirow}
\usepackage{makecell}
\usepackage{color}
\usepackage{booktabs}
\usepackage{epsfig}
\usepackage{graphicx}
\usepackage{amsmath}
\usepackage{amssymb}
\usepackage{bm}
\usepackage{colortbl}
\usepackage{bbding}
\usepackage{pifont}

\definecolor{graycolor}{rgb}{0.95,0.95,0.95}

\def\ie{\emph{i.e.}} 

\def\eg{\emph{e.g.}}

\def\etal{\emph{et al. }}

\newcommand{\blue}[1]{\textcolor{cvprblue}{#1}}

\def\paperID{} 
\def\confName{CVPR}
\def\confYear{2026}

\title{Enhancing Unregistered Hyperspectral Image Super-Resolution via \\ Unmixing-based Abundance Fusion Learning}

\author{Yingkai Zhang\textsuperscript{1} \qquad Tao Zhang\textsuperscript{2} \qquad Jing Nie\textsuperscript{1} \qquad Ying Fu\textsuperscript{1$\dagger$}\\
\textsuperscript{1}Beijing Institute of Technology \quad \textsuperscript{2}Hangzhou Dianzi University \\
{\tt\small zhangyingkai@bit.edu.cn \quad tzhang@hdu.edu.cn \quad \{3420235028, fuying@bit.edu.cn\}}}

\begin{document}
\maketitle

{
\renewcommand{\thefootnote}{}
\footnotetext[5]{$\dagger$ Corresponding author.} 
}

\begin{abstract}
Unregistered hyperspectral image (HSI) super-resolution (SR) typically aims to enhance a low-resolution HSI using an unregistered high-resolution reference image.
In this paper, we propose an unmixing-based fusion framework that decouples spatial-spectral information to simultaneously mitigate the impact of unregistered fusion and enhance the learnability of SR models.
Specifically, we first utilize singular value decomposition for initial spectral unmixing, preserving the original endmembers while dedicating the subsequent network to enhancing the initial abundance map.
To leverage the spatial texture of the unregistered reference, we introduce a coarse-to-fine deformable aggregation module, which first estimates a pixel-level flow and a similarity map using a coarse pyramid predictor. It further performs fine sub-pixel refinement to achieve deformable aggregation of the reference features. 
The aggregative features are then refined via a series of spatial-channel abundance cross-attention blocks. 
Furthermore, a spatial-channel modulated fusion module is presented to merge encoder-decoder features using dynamic gating weights, yielding a high-quality, high-resolution HSI. 
Experimental results on simulated and real datasets confirm that our proposed method achieves state-of-the-art super-resolution performance. The code will be available at \url{https://github.com/yingkai-zhang/UAFL}.
\end{abstract}

\begin{figure}
\begin{center}
\includegraphics[width=1\linewidth]{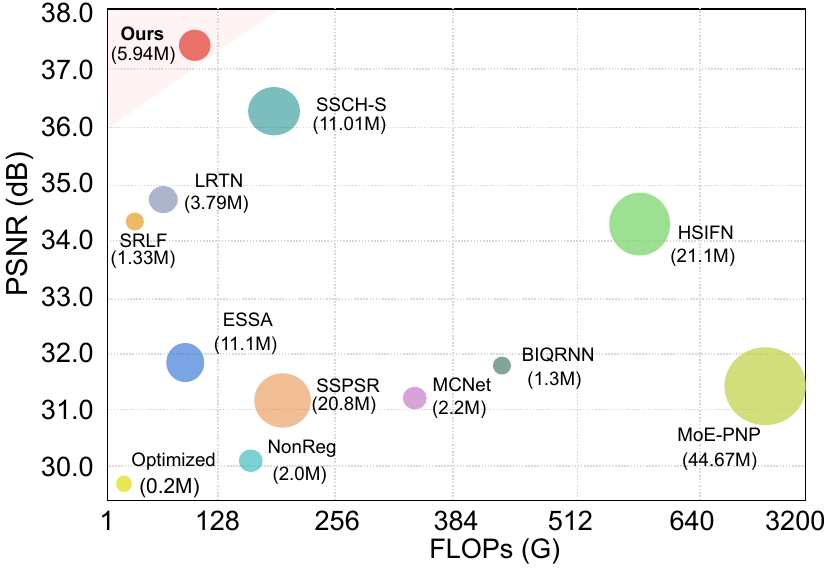}
\end{center}
	\vspace{-5mm}
   \caption{Comparison of performance (PSNR), computational cost (FLOPs), and memory cost (Params). The vertical axis represents PSNR (dB), while the horizontal axis indicates FLOPs. The size of each bubble corresponds to the number of parameters.}
   \vspace{-5mm}
\label{fig:params_flops}
\end{figure}

\section{Introduction}
\label{sec:intro}

Hyperspectral images (HSIs) provide extensive spectral information that has driven progress in a variety of fields~\cite{thenkabail2016hyperspectral, ram2024systematic, pan2003face}. However, HSI sensors usually face a well-known trade-off between spatial and spectral resolution. As a result, HSIs often have high spectral accuracy but limited spatial detail. Therefore, hyperspectral Super-Resolution (SR) has become a vital area of research.

Single HSI SR methods~\cite{zhang2023essaformer, fu2022coded, liang2023blind} have shown some effectiveness, but they are inherently limited by the finite information available in a single input.
Consequently, there has been a growing focus on reference-based registered HSI SR~\cite{liu2025low, liu2025selective}, which utilizes an auxiliary high-resolution (HR) reference image to augment the spatial resolution of a low-resolution (LR) HSI.
While effective, they are limited by the restrictive assumption that the LR HSI and HR reference image are perfectly aligned. This is seldom the case in practical applications, where misalignments inevitably arise from platform vibrations, perspective shifts, or differences in sensor acquisition times~\cite{ying2022unaligned}. 
This challenge has led to the development of methods~\cite{zheng2022nonregsrnet, lai2024hyperspectral} for unregistered HSI SR, which aim to perform alignment and fusion jointly. 

\begin{figure}
\begin{center}
\includegraphics[width=1\linewidth]{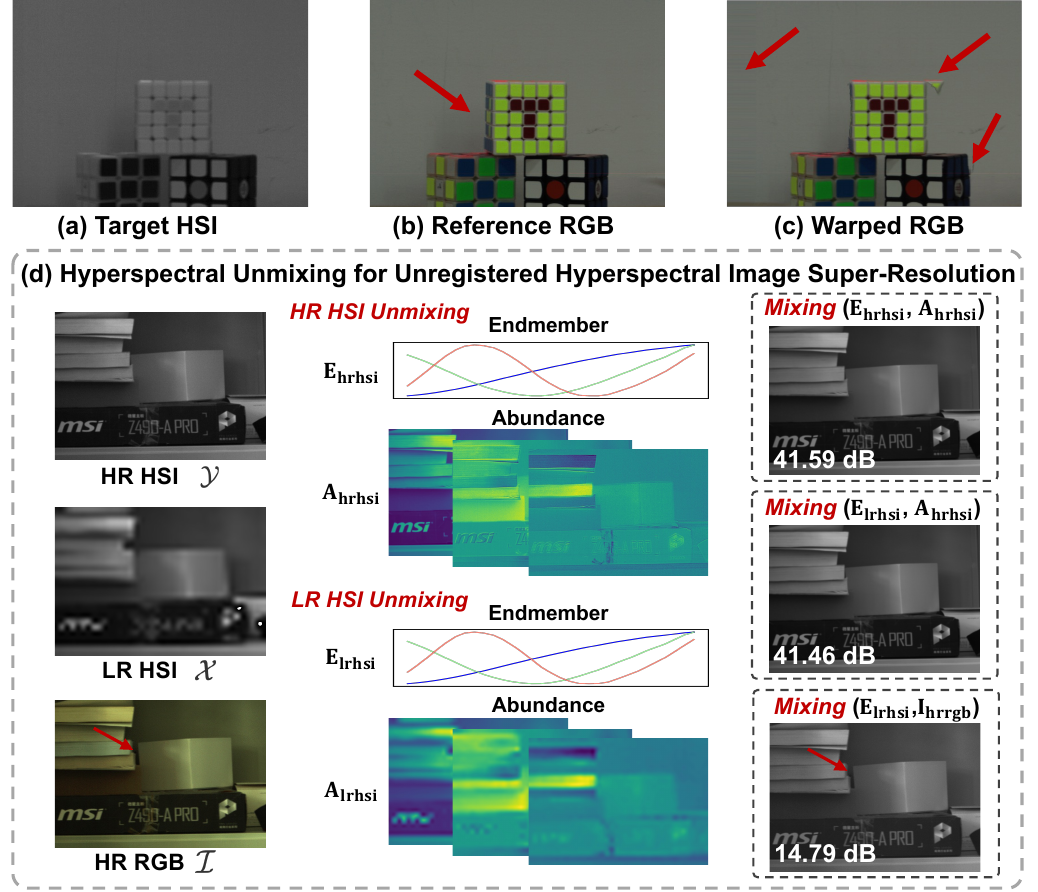}
\end{center}
	\vspace{-4mm}
   \caption{The motivation. (c) The explicit alignment within the SSCH~\cite{zhang2025unaligned} is prone to introducing spatial distortions and textural artifacts into the warped image. (d) The result validates the principle that leveraging a well-aligned, high-fidelity abundance enables the reconstruction of the HR HSI from its LR counterpart.}
   \vspace{-5mm}
\label{fig:Motivation}
\end{figure}

Some works~\cite{lai2024hyperspectral, zhang2025unaligned} typically follow a two-stage pipeline where a pre-trained module (\eg, RAFT~\cite{teed2020raft}) first explicitly aligns the reference image, which is then fused by a subsequent spatial-spectral coupled network. Nevertheless, these approaches have two primary challenges. 
\textbf{First}, the reliance on coupled spatial-spectral fusion constrains the learning capacity of networks (Fig.~\ref{fig:params_flops}). 
\textbf{Second}, the explicit alignment step often introduces textural distortions and artifacts (Fig.~\ref{fig:Motivation}\blue{(a-c)}) due to inherent disparities in resolution and data distribution. 

The inherent low-rank property of HSIs~\cite{peng2021low, li2024latent}, stemming from strong spectral correlations, is a cornerstone of restoration tasks~\cite{he2019non, wei2019low} and underpins spectral unmixing~\cite{heylen2014review}.
As discussed in the work~\cite{fu2020simultaneous}, unmixing-based methods~\cite{kawakami2011high, akhtar2014sparse} are robust to geometric misalignment, while others~\cite{dian2018deep, dong2016hyperspectral} are more sensitive, employing alignment constraint.
Inspired by this, we provide an analysis of spectral unmixing for unregistered HSI SR in Fig.~\ref{fig:Motivation}\blue{(d)}. 
It can be seen that combining LR HSI endmembers with well-structured HR abundances yields superior results, in contrast to direct mixing with an unregistered reference, which produces a visually plausible image but suffers from poor quantitative performance due to real misalignment.
This decoupling strategy transforms the complex problem into a more specific learning objective, which simplifies the optimization process for the network.

In this paper, we propose an unmixing-based fusion framework that leverages deformable aggregation and spatial-channel modulated fusion to guide HSI super-resolution. 
Specifically, our approach reframes the task from direct fusion of spatial-spectral to learning residual abundance maps.
We first utilize singular value decomposition to decouple HSI into endmembers and abundance, which is further enhanced with an unregistered reference image.
Next, we present a coarse-to-fine deformable module that performs implicit feature aggregation by first estimating a pixel-level flow and similarity map with a coarse pyramid predictor, and then executing a fine sub-pixel refinement using deformable convolution for an effective aggregation of reference features.
The resulting features are then refined by a spatial-channel abundance cross-attention mechanism to enhance their representation capability. 
Besides, a spatial-channel modulated fusion module in the decoder generates dynamic gating weights to optimally merge encoder-decoder features, ensuring a high-fidelity reconstruction of both spatial details and spectral signatures.
The main contributions of our work are summarized as follows:
\begin{itemize}
    \item We propose an unmixing-based fusion framework for unregistered HSI SR by decoupling spatial-spectral information to mitigate the impact of unregistered fusion and enhance the learnability of SR model.
    \item We introduce a coarse-to-fine deformable aggregation module with spatial-channel abundance cross-attention for robust feature aggregation and enhanced representational capability.
    \item We design a spatial-channel modulated fusion module that generates dynamic gating weights to dynamically merge encoder-decoder features. This ensures a high-fidelity reconstruction of both fine spatial details and spectral signatures in the final HSI.
\end{itemize}

\section{Related Work}

In this section, we review the reference-based super-resolution, including registered HSI super-resolution and unregistered HSI super-resolution.

\subsection{Registered HSI Super-Resolution}

Registered HSI super-resolution (SR) leverages an HR reference image to provide spatial details. The majority of these methods follow the strict assumption that the LR HSI and the HR reference image are perfectly registered. 
One line of research focuses on optimization techniques with hand-crafted priors~\cite{akhtar2014sparse, akhtar2015bayesian, kawakami2011high, lanaras2015hyperspectral, heylen2014review, zhu2025self}.
Although these methods can be robust to geometric misalignment to a certain extent, they struggle with complex, real-world scenarios due to the dependence on assumed priors.

In recent years, various deep learning methods, including CNNs and Transformers~\cite{zhang2025real, zhang2026supervise, wang2025erienet, zhang2024deep, tian2023transformer, li2025fcdfusion, li2025noise, gao2025grayscale, zhang2022guided, zhang2024rgb, zhang2025unaligned, chen2026degraded} have been proposed.
Another line of research based on these~\cite{hu2022hyperspectral, fu2019hyperspectral, zhang2024rgb, liu2025selective, hou2025bidomain, hou2025binarized, liu2025physicsinformed}, learn the mapping from LR HSI and HR reference to HR HSI directly from data, removing the need for hand-crafted priors. 
However, the performance of these methods hinges on the assumption of perfectly aligned data. This idealized condition is difficult to satisfy in real-world scenarios, which seriously undermines the practical applicability of these registered fusion approaches.

\begin{figure*}[t]
   \begin{center}
   \includegraphics[width=1\linewidth]{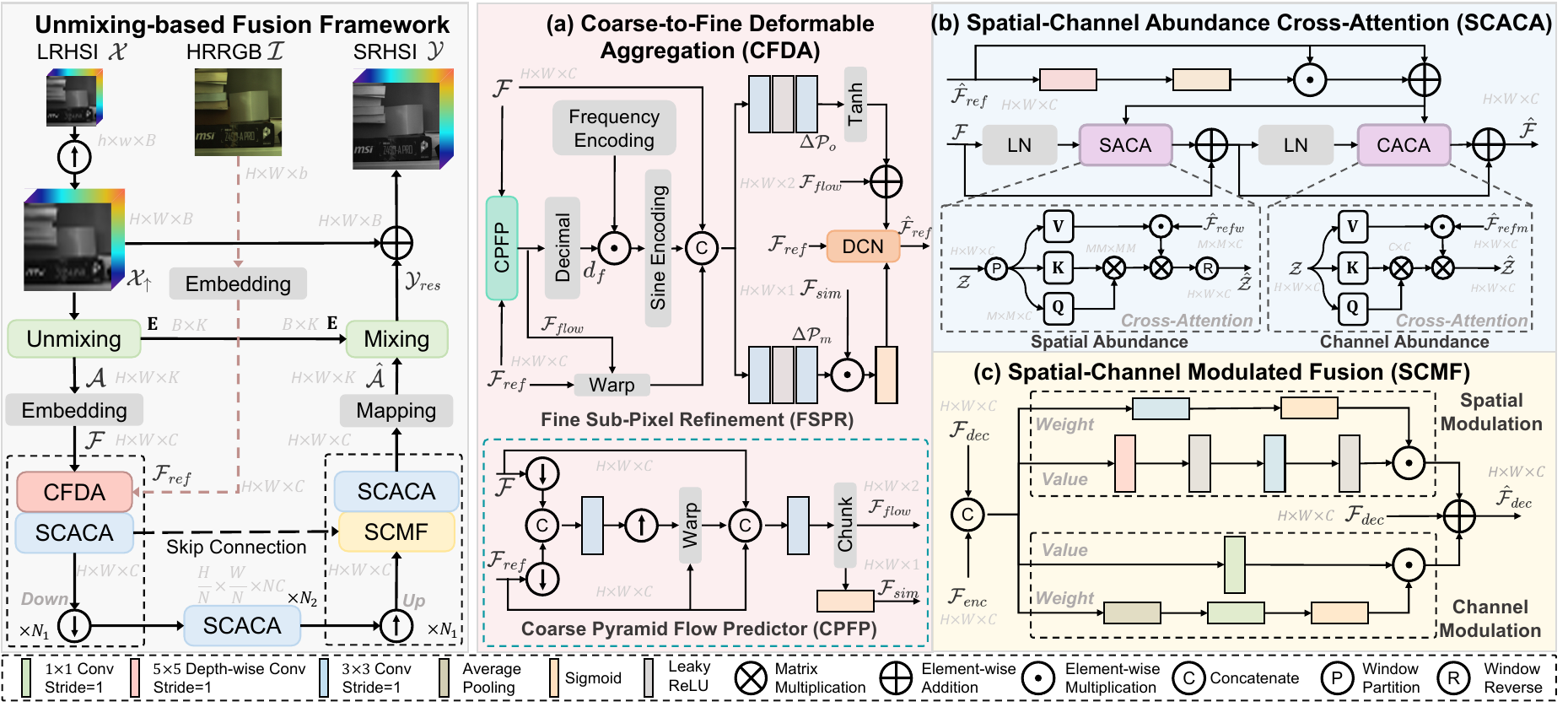}
   \end{center}
   \vspace{-4mm}
	\caption{The overall architecture. We adopt a multi-scale encoder-decoder based on unmixing with several key modules: (a) Coarse-to-Fine Deformable Aggregation module, (b) Spatial-Channel Abundance Cross-Attention module, and (c) Spatial-Channel Modulation Fusion module to super-resolve high-quality HR HSI.}
	\vspace{-4mm}
   \label{fig:framework}
   \end{figure*}

\subsection{Unregistered HSI Super-Resolution}

To mitigate the reliance on strict alignment priors and enhance the practical applicability, some works begin to study unregistered HSI SR~\cite{fu2020simultaneous, nie2020unsupervised, qu2022unsupervised, zheng2022nonregsrnet, zhou2019integrated, ying2022unaligned}. These approaches typically train on unregistered data simulated through rigid or non-rigid transformations. 
The predominant strategy involves a two-stage pipeline where an explicit transformation is first estimated and then applied. For instance, some methods conduct rigid or non-rigid transformation through optimization~\cite{fu2020simultaneous}, spatial transformer network~\cite{nie2020unsupervised}, or optical flow model~\cite{lai2024hyperspectral}.
Zhang \etal~\cite{zhang2025unaligned} first utilize a pre-trained flow model, and then finetune the feature in a post-process fusion network. 
The dominant approaches are hindered by the fact that explicit pre-alignment often introduces textural artifacts, while integrated methods~\cite{zheng2022nonregsrnet, qu2024principle} suffer from the learning constraints imposed by coupled spatial-spectral data.
Therefore, we propose an unmixing-based fusion network that implicitly aggregates features, complemented by cross-attention and modulated fusion, to mitigate the impact of unregistered fusion and enhance representational capability.

\section{Unmixing-based Fusion Framework}

In this section, we first introduce the motivation and formulation of our framework in \ref{sec:MF}. Then, we present the details of key modules in \ref{sec:CFDA}, \ref{sec:SCACA}, and \ref{sec:SCMF}.

\subsection{Motivation and Overall Framework}
\label{sec:MF}

\noindent{\textbf{{Motivation.}}
Spectral unmixing, enabled by the inherent low-rank property of HSIs~\cite{heylen2014review, fu2020simultaneous, peng2021low, xue2021spatial}, can mitigate the adverse effects of unregistered fusion, as discussed in work~\cite{fu2020simultaneous}. Inspired by this, as shown in Fig.~\ref{fig:Motivation}\blue{(d)}, we provide unmixing analysis on HR HSI, LR HSI, and HR unregistered RGB.
The matrix and tensor are defined as $\mathbf{X}$ and $\mathcal{X}$, respectively.
A high-quality HSI can be reconstructed by fusing endmembers from the LR HSI with abundances from an HR source, \ie, Mixing($\mathbf{E}_{lrhsi},\mathbf{A}_{hrhsi}$). This allows us to reframe the fusion task from learning in a complex, coupled spatial-spectral domain to a more tractable problem: learning to enhance the decoupled abundance maps. 
While a reference RGB image can provide the necessary spatial structure for these abundances, direct mix by $\mathbf{E}_{lrhsi}$ and $\mathbf{I}_{hrrgb}$ is quantitatively poor due to real misalignment, even if it appears visually sharp (Fig.~\ref{fig:Motivation}\blue{(d)}). 
Consequently, to effectively aggregate unregistered reference features, we propose a coarse-to-fine deformable aggregation module. By operating on deep features rather than pixels, our module with cross-attention blocks learns robust and clear representations and makes more effective use of the unregistered reference to guide the abundance map enhancement.

The architecture of our framework is illustrated in Fig.~\ref{fig:framework}. Given an LR HSI $\mathcal{X} \in \mathbb{R}^{h \times w \times B}$ and an unregistered HR RGB $\mathcal{I} \in \mathbb{R}^{H \times W \times b}$, our goal is to reconstruct an HR HSI $\mathcal{Y} \in \mathbb{R}^{H \times W \times B}$, where $H$, $W$, and $B$ denote the high-resolution height, width, and number of spectral bands.

\noindent{\textbf{Unmixing.}}
To initialize the unmixing process, we first upsample the LR HSI $\mathcal{X}$ to the target spatial dimensions, HSI $\mathcal{X}_{\uparrow} \in \mathbb{R}^{H \times W \times B}$. Subsequently, we apply singular value decomposition to decouple $\mathcal{X}_{\uparrow}$ by $\mathbf{X}_{\uparrow} = \mathbf{U}\mathbf{S}\mathbf{V}^\mathrm{T}$, where $\mathbf{U} \in \mathbb{R}^{B \times B}$, $\mathbf{S} \in \mathbb{R}^{B \times HW}$, $\mathbf{V} \in \mathbb{R}^{HW \times HW}$, and $\mathbf{X}_{\uparrow}$ denotes that $\mathcal{X}_{\uparrow}$ is reshaped to $\mathbb{R}^{B \times HW}$. The pre-estimated endmember matrix $\mathbf{E} \in \mathbb{R}^{B \times K}$ is formed by taking the first $K$ left singular vectors from $\mathbf{U}$, where $K$ is the number of endmembers. The abundance $\mathbf{A} \in \mathbb{R}^{K \times HW}$ is obtained by:
\begin{equation}
    \mathbf{A} = \mathbf{E}^\mathrm{T}\mathbf{X}_{\uparrow},
\end{equation}
where $\mathbf{A}$ is an initial abundance map, which is then reshaped to $\mathcal{A} \in \mathbb{R}^{H \times W \times K}$ for subsequent learning residual abundance maps with the unregistered reference $\mathcal{I}$ by:
\begin{equation}
    \hat{\mathcal{A}} = f(\mathcal{A}, \mathcal{I}\vert\theta),
\end{equation}
where $\hat{\mathcal{A}}$ is the learned residual abundance map, $f(\cdot\vert\theta)$ denotes our network, and $\theta$ represents learnable parameters. 

\noindent{\textbf{Mixing.}}
We then reconstruct a residual HSI component by mixing $\hat{\mathcal{A}}$ with the initial endmembers $\mathbf{E}$:
\begin{equation}
    \mathbf{Y}_{res} = \mathbf{E}\hat{\mathbf{A}},
\end{equation}
where $\hat{\mathbf{A}} \in \mathbb{R}^{K \times HW}$ and $\mathbf{Y}_{res} \in \mathbb{R}^{B \times HW}$. 
The super-resolved HR HSI $\mathcal{Y}$ is obtained by adding this learned residual component to the initial upsampled HSI:
\begin{equation}
    \mathcal{Y} = \mathcal{Y}_{res} + \mathcal{X}_{\uparrow},
\end{equation}
where $\mathcal{Y}_{res}$ denotes that $\mathbf{Y}_{res}$ is reshaped to $\mathbb{R}^{H \times W \times B}$. 

\subsection{Coarse-to-Fine Deformable Aggregation}
\label{sec:CFDA}

To effectively leverage the spatial information from the unregistered reference image $\mathcal{I}$ to enhance the initial abundance map $\mathcal{A}$, we introduce a Coarse-to-Fine Deformable Aggregation (CFDA) module (Fig.~\ref{fig:framework}\blue{(a)}). 
The core of this module is the deformable convolution guided by a dynamically predicted prior flow, which is progressively refined. 
We first extract deep features from both the initial abundance map and the reference image, denoted as $\mathcal{F} \in \mathbb{R}^{H \times W \times C}$ and $\mathcal{F}_{ref} \in \mathbb{R}^{H \times W \times C}$, respectively. 

\noindent{\textbf{Coarse Pyramid Flow Predictor}}.
For the initial coarse motion estimation, we utilize a lightweight pyramid predictor to compute a preliminary flow field.
The feature $\mathcal{F}$ and reference feature $\mathcal{F}_{ref}$ are first downsampled as $\mathcal{F}_{\downarrow}$ and $\mathcal{F}_{ref\downarrow}$, and a convolutional layer predicts a low-resolution flow field. This flow is then upsampled to the original resolution to serve as a coarse motion prior:
\begin{equation}
    \mathcal{C}_{flow} = \text{Up}(\text{Conv}_{3\times3}(\mathcal{F}_{\downarrow}, \mathcal{F}_{ref\downarrow})),
\end{equation}
where $\text{Up}(\cdot)$ denotes bilinear upsampling, $\text{Conv}_{3\times3}(\cdot,\cdot)$ denotes the convolution with kernel size $3\times3$ for concatenation of inputs. 
Then, $\mathcal{F}_{ref}$ is warped using this coarse flow. The warped feature is then concatenated with $\mathcal{F}$ and fed into another convolutional layer to predict a residual flow $\Delta \mathcal{C}_{flow}$ and the similarity map $\mathcal{F}'_{sim}$. The final prior flow $\mathcal{F}_{flow} \in \mathbb{R}^{H \times W \times 2}$ is the sum of the coarse and residual components, and prior similarity map $\mathcal{F}_{sim} \in \mathbb{R}^{H \times W \times 1}$, which indicates regions of high confidence, is obtained by applying a sigmoid function:
\begin{equation}
	\begin{aligned}
		\mathcal{F}_{flow} &= \mathcal{C}_{flow} + \Delta \mathcal{C}_{flow}, \\
		\mathcal{F}_{sim} &= \text{Sigmoid}(\mathcal{F}'_{sim}).
	\end{aligned}
\end{equation}

\noindent{\textbf{Fine Sub-Pixel Refinement}}.
The predicted flow provides a coarse displacement but lacks fine-grained precision. To further enhance this, and inspired by the work~\cite{xu2024enhancing}, we encode sub-pixel displacements, providing a detailed prior that enables the network to capture fine local details at each sampling location.
We first extract the decimal component $\mathbf{d}_f \in \mathbb{R}^{H \times W \times 2}$ of the flow $\mathcal{F}_{flow}$. We follow previous works~\cite{xu2024enhancing, mildenhall2021nerf} and adopt a frequency positional encoding:
\begin{equation}
	\gamma(\mathbf{d}_f) = [\omega\mathbf{d}_f, \omega^2\mathbf{d}_f, \cdot\cdot\cdot, \omega^{N-1}\mathbf{d}_f],
\end{equation}
where, $\gamma(\mathbf{d}_f) \in \mathbb{R}^{H \times W \times 2N}$, $\omega$ is the angular speed and $N$ denotes the frequency bands. We then obtain a high-dimensional feature positional encoding, $\mathcal{F}_{pe} \in \mathbb{R}^{H \times W \times 4N}$ that represents fine-grained sub-pixel positional information by concatenating a series of sine and cosine functions with varying frequencies:
\begin{equation}
	\mathcal{F}_{pe} = \text{Concat}[\text{Sin}(\gamma(\mathbf{d}_f)), \text{Cos}(\gamma(\mathbf{d}_f))].
\end{equation}
Then, a refinement network takes the concatenation of the target feature, the warped reference feature, and the positional features to predict a residual offset and mask:
\begin{equation}
	\Delta \mathcal{P} = f_{re}(\text{Concat}[\mathcal{F}, \text{Warp}(\mathcal{F}_{ref}, \mathcal{F}_{flow}), \mathcal{F}_{pe}]),
\end{equation}
where $\text{Warp}(\cdot)$ is the warping operation. $\Delta \mathcal{P}$ can be chunked to $\Delta \mathcal{P}_{o}$ and $\Delta \mathcal{P}_{m}$, which perform fine-grained corrections on the prior coarse flow.
The final offset $\mathcal{O}$ and modulation mask $\mathcal{M}$ for deformable can be obtained by:
\begin{equation}
	\begin{aligned}
		\mathcal{O} &= \mathcal{F}_{flow} + \text{Tanh}(\Delta \mathcal{P}_{o}), \\
		\mathcal{M} &= \text{Sigmoid}(\mathcal{F}_{sim}\odot \Delta \mathcal{P}_{m}),
	\end{aligned}
\end{equation}
where $\odot$ denotes element-wise multiplication. Finally, the aggregative reference feature $\hat{\mathcal{F}}_{ref}$ is obtained by applying the modulated deformable convolution to the original reference feature $\mathcal{F}_{ref}$ using the offsets $\mathcal{O}$ and masks $\mathcal{M}$.

\subsection{Spatial-Channel Abundance Cross-Attention}
\label{sec:SCACA}

After obtaining the aggregative features from the CFDA module, we introduce a Spatial-Channel Abundance Cross-Attention (SCACA) block to further refine the abundance map features, as shown in Fig.~\ref{fig:framework}\blue{(b)}.

We first modulate the aggregative reference feature $\hat{\mathcal{F}}_{ref}$ by a lightweight self-modulated network:
\begin{equation}
	\hat{\mathcal{F}}_{refm} = \hat{\mathcal{F}}_{ref} + \hat{\mathcal{F}}_{ref} \odot \text{Sigmoid}(\text{Conv}_{5\times5}(\hat{\mathcal{F}}_{ref})),
\end{equation}
where $\text{Conv}_{5\times5}$ denotes the $5\times5$ depth-wise convolution.
Inspired by the work~\cite{yao2024specat}, we then utilize hierarchical cross-attention (sequentially spatial and channel cross-attention) to restore their fine-grained details.

\noindent{\textbf{Spatial Abundance Cross-Attention (SACA)}}. 
To enhance the spatial structure of the abundance map, we employ a window-based cross-attention.
The abundance feature $\mathcal{F}$ is first processed by a LayerNorm (LN) layer, then interacted with the modulated aggregative reference feature $\hat{\mathcal{F}}_{refm}$ by the cross-attention block:
\begin{equation}
	\mathcal{F}' = \mathcal{F}+\text{FFN}(\text{SACA}(\text{LN}(\mathcal{F}), \hat{\mathcal{F}}_{refm})),
\end{equation}
where $\text{FFN}(\cdot)$ consists of several convolutional layers and GELU activation layers. For attention block, the input abundance feature $\mathcal{Z}$ is first conducted a window partition operation into $\mathcal{Z}_w \in \mathbb{R}^{M\times M\times C}$, where $M$ is the window size. Similarly, $\hat{\mathcal{F}}_{refm}$ is divided into non-overlapping patches, where single patch is $\hat{\mathcal{F}}_{refw} \in \mathbb{R}^{M\times M\times C}$.
Thus, the Query ($\mathcal{Q}$), Key ($\mathcal{K}$), and Value ($\mathcal{V}$) matrices are all derived from $\mathcal{Z}_w$. The cross-attention mechanism is introduced by modulating the Value vector with the reference feature before the final attention aggregation.
The core operation within a window can be summarized as:
\begin{equation}
	\begin{aligned}
		\mathcal{V}_{mod} = \mathcal{V} \odot \text{Reshape}(\hat{\mathcal{F}}_{refw}), \\
		\hat{\mathcal{Z}} = \text{Softmax}\left(\frac{\mathcal{Q}\mathcal{K}^T}{\sqrt{d_k}} + \mathcal{B}\right)\mathcal{V}_{mod},
	\end{aligned}
\end{equation}
where $\mathcal{B}$ is the learnable relative position bias. This allows the model to learn fine-grained spatial correspondence by leveraging the structural information from the reference feature to guide the refinement of the abundance map.

\noindent{\textbf{Channel Abundance Cross-Attention (CACA)}}.
Complementary to the spatial attention, the CACA block focuses on refining the channel-wise signatures of the abundance map, dynamically enhancing salient spectral signatures while attenuating those that are irrelevant.
Similar to the spatial attention, the $\mathcal{Q}$, $\mathcal{K}$, and $\mathcal{V}$ matrices are derived from the intermediate abundance feature $\mathcal{F}'$. The channel cross-attention is realized by modulating the Value vector:
\begin{equation}
    \mathcal{V}_{mod} = \mathcal{V} \odot \hat{\mathcal{F}}_{refm}.
\end{equation}
This operation adaptively recalibrates the channel-wise responses of the abundance features based on the channel characteristics of the aggregative reference features.
By combining these two mechanisms, the SCACA module effectively fuses multi-modal information, leading to a refined abundance map.

\subsection{Spatial-Channel Modulated Fusion}
\label{sec:SCMF}

To effectively merge the hierarchical features from the encoder-decoder architecture and ensure a high-fidelity reconstruction, we propose a Spatial-Channel Modulated Fusion (SCMF) module. This module is designed to adaptively fuse features from a decoder stage, $\mathcal{F}_{dec}$, with corresponding features from an encoder stage, $\mathcal{F}_{enc}$. 
The core principle is to decouple the fusion process into two parallel, complementary pathways, one focusing on spatial modulation and the other on channel modulation.
The two features $\mathcal{F}_{enc}$ and $\mathcal{F}_{dec}$ are first concatenated along the channel dimension to form a unified feature map, $\mathcal{F}_{cat}$, which serves as the input for modulated fusion.

\noindent\textbf{{Spatial Modulation}}.
This modulation is designed to emphasize or suppress spatial details based on local context. It consists of a spatial value branch and a spatial gating branch. The value branch, composed of depth-wise convolutions and Leaky ReLU, processes the concatenated feature to generate a spatially-rich representation, $\mathcal{V}_{spa}$.
Concurrently, the spatial gating branch generates a pixel-wise gating weight, $\mathcal{W}_{spa}$. The weight is produced by a convolution followed by a sigmoid activation function, which assigns an importance weight to each spatial location:
\begin{equation}
    \mathcal{M}_{spa} = \text{Sigmoid}(\text{Conv}_{3 \times 3}(\mathcal{F}_{cat})).
\end{equation}
The output of this modulation is the element-wise product of the value and the gating weight, $\mathcal{F}_{spa} = \mathcal{V}_{spa} \odot \mathcal{M}_{spa}$, where $\mathcal{M}_{spa}$ is broadcast across the channel dimension.

\noindent\textbf{{Channel Modulation}}.
Complementary to the spatial modulation, the channel modulation focuses on adaptively recalibrating channel-wise feature responses. It also consists of a channel value branch and a channel gating branch. The value branch uses a $1 \times 1$ convolution to produce a channel-mixed representation, $\mathcal{V}_{spe}$.
The channel gating branch first aggregates global spatial information into a channel descriptor using Global Average Pooling (GAP). This descriptor is then transformed by a $1 \times 1$ convolution to compute channel-wise attention weights, which are normalized by a sigmoid function:
\begin{equation}
    \mathcal{M}_{spe} = \text{Sigmoid}(\text{Conv}_{1 \times 1}(\text{GAP}(\mathcal{F}_{cat}))).
\end{equation}
The resulting attention vector $\mathcal{M}_{spe} \in \mathbb{R}^{1 \times 1 \times C}$ is used to modulate the channel value feature, yielding the output $\mathcal{F}_{spe} = \mathcal{V}_{spe} \odot \mathcal{M}_{spe}$.

To dynamically balance the contributions of spatial and channel information during the fusion process, leading to a more robust and high-fidelity reconstruction, the outputs of the two modulations are fused via element-wise addition. Besides, to ensure stable training and preserve information from the decoder stream, this fused result is added to the original decoder feature $\mathcal{F}_{dec}$ in a residual connection:
\begin{equation}
    \hat{\mathcal{F}}_{dec} = (\mathcal{F}_{spa} + \mathcal{F}_{spe}) + \mathcal{F}_{dec}.
\end{equation}

\begin{figure*}
 \centering
\small
\hspace{1mm}
 \setlength{\tabcolsep}{0.1cm}
\begin{minipage}[l]{0.22\linewidth}
\begin{flushleft}
 \vspace{-0.1mm}
  \hspace{-3mm}
\includegraphics[width=1.045\linewidth]{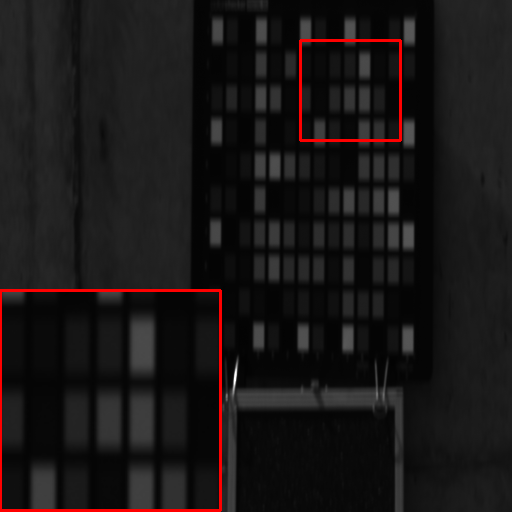}\\
   \begin{tabular}[c]{@{}c@{}}\hspace{2mm}Ground Truth (ICVL~\cite{arad2016sparse})\end{tabular}
  \end{flushleft}
 \end{minipage}
 \hspace{-1mm}
 \begin{minipage}[t]{0.5\linewidth}
  \begin{tabular}{cccccc}
   \includegraphics[width=0.2\linewidth]{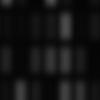}
  & \includegraphics[width=0.2\linewidth]{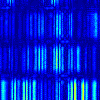}
  & \includegraphics[width=0.2\linewidth]{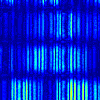}
  & \includegraphics[width=0.2\linewidth]{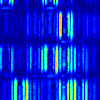}
  & \includegraphics[width=0.2\linewidth]{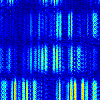}
  & \includegraphics[width=0.2\linewidth]{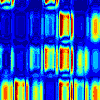}
   \\
	Input ($\times 4$) & SSPSR~\cite{jiang2020learning}  & MCNet~\cite{li2020mixed} & BiQRNN~\cite{fu2021bidirectional} & ESSA~\cite{zhang2023essaformer}  & NonReg~\cite{zheng2022nonregsrnet} \\

 \includegraphics[width=0.2\linewidth]{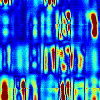}
  &\includegraphics[width=0.2\linewidth]{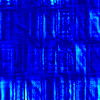}
  &\includegraphics[width=0.2\linewidth]{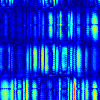}
  &\includegraphics[width=0.2\linewidth]{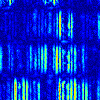}
  &\includegraphics[width=0.2\linewidth]{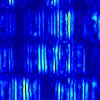}
  &\includegraphics[width=0.2\linewidth]{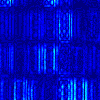}
  
  \\      
	MoE-PNP~\cite{qu2024principle} & HSIFN~\cite{lai2024hyperspectral} & LRTN~\cite{liu2025low}  & SRLF~\cite{liu2025selective} & SSCH-S~\cite{zhang2025unaligned} & Ours
  \end{tabular}
 \end{minipage}
 \hspace{22mm}
 \begin{minipage}[l]{0.1\linewidth}
 \begin{flushright}
    \vspace{-2mm}
     \includegraphics[width=0.54\linewidth]{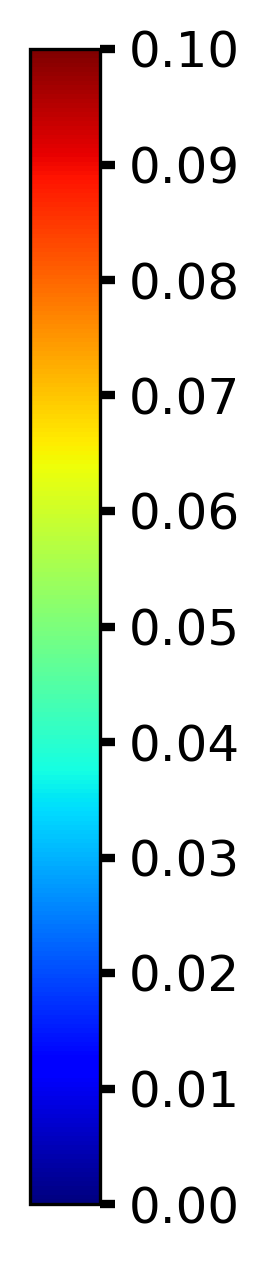}
 \end{flushright}
 \end{minipage}

 \caption{Visual comparison on the simulated dataset, ICVL~\cite{arad2016sparse}. The ground truth and input HSIs are shown with band 20. The results of different methods under the scale factor ($\times4$) on the ICVL dataset with the error maps.}
 \label{fig:icvl visual results} 
\end{figure*}

\section{Experiments}

In this section, we first introduce the experimental settings. Then, we compare our approach with several state-of-the-art methods on simulated and real data. We further conduct ablation studies to validate the effectiveness of our modules.

\subsection{Experimental Settings}

\noindent\textbf{{Datasets and Metrics}}.
We utilize two distinct datasets, ICVL~\cite{arad2016sparse} and REAL~\cite{lai2024hyperspectral}.
The ICVL dataset contains 201 pairs of images with a resolution of $1300\times1392$, which are cut into a size of 512 by the center in the experiment. The RGB reference image is generated by the spectral response function of Nikon D700 as~\cite{akhtar2015bayesian}. The misregistration is achieved by adjusting the pose of the virtual camera to control the generation of new viewpoints~\cite{liang_2023_ICCV,zhang2025unaligned}.
The REAL dataset comprises 60 paired, unregistered HSI images of both indoor and outdoor scenes, captured by a dual-camera system equipped with SOC710-VP hyperspectral camera.
Following the works~\cite{lai2024hyperspectral, zhang2025unaligned}, we choose 20 pairs for testing and the rest for training in ICVL dataset, and 10 pairs for testing and the rest for training in REAL dataset.
To evaluate the performance, we employ three metrics, \ie, peak signal-to-noise ratio (PSNR), structural similarity index (SSIM), and spectral angle mapping (SAM).

\noindent\textbf{{Implementation Details}}.
Following the work~\cite{zhang2025unaligned}, we utilize the Gaussian kernel ($\mu=8, \sigma=3$) to blur the HSI and downsample with the scale factor (\eg, $\times4$, $\times8$, $\times16$) to acquire the LR HSI.
We develop our model using Pytorch~\cite{paszke2019pytorch} with a batch size of 1, trained on a single Nvidia RTX 4090 GPU. 
The model is trained via the AdamW~\cite{loshchilov2017decoupled} optimizer with the weight decay rate of $5\times10^{-5}$ and learning rate of $1\times10^{-5}$ to minimize the $\bm{L}_1$ loss function for 150 epochs in ICVL dataset and 300 epochs in REAL dataset.
All methods are trained with the same settings as ours for fair comparison.

\subsection{Main Results}

\noindent\textbf{{Competing Methods}}.
We compare the proposed method with several state-of-the-art (SOTA) SR methods, including single HSI SR methods (SSPSR~\cite{jiang2020learning}, MCNet~\cite{li2020mixed}, BiQRNN~\cite{fu2021bidirectional}, and ESSA~\cite{zhang2023essaformer}) and reference-based HSI SR methods (Optimized~\cite{fu2019hyperspectral}, NonReg~\cite{zheng2022nonregsrnet}, MoE-PNP~\cite{qu2024principle}, HSIFN~\cite{lai2024hyperspectral}, LRTN~\cite{liu2025low}, SRLF~\cite{liu2025selective}, and SSCH~\cite{zhang2025unaligned}).

\begin{table}[t]
    \centering
    \small
    \caption{Quantitative comparison under scale factor of $\times4$ on the simulated dataset, ICVL~\cite{arad2016sparse}. PSNR is in dB. The best and second best results are highlighted in \textbf{bold} and \underline{underline}, respectively.}
    \label{tab:result_on_icvl}
    \renewcommand\arraystretch{0.9} 
    \setlength{\tabcolsep}{6pt} 
    \small
    \begin{tabular}{l|c|ccc}
        \toprule[1.2pt]
        \textbf{Method} & \textbf{Venue} & PSNR$\uparrow$ & SSIM$\uparrow$ & SAM$\downarrow$ \\
        \midrule
        Bicubic & - & 35.33 & 0.960 & \underline{0.028} \\
        SSPSR~\cite{jiang2020learning} & TCI'20 & 40.19 & 0.982 & 0.033 \\
        MCNet~\cite{li2020mixed} & RS'20 & 39.90 & 0.980 & 0.033 \\
        BiQRNN~\cite{fu2021bidirectional} & JSTAR'21 & 37.85 & 0.974 & 0.039 \\
        ESSA~\cite{zhang2023essaformer} & ICCV'23 & 38.41 & 0.977 & 0.057 \\
        Optimized~\cite{fu2019hyperspectral} & CVPR'19 & 25.35 & 0.801 & 0.026 \\
        NonReg~\cite{zheng2022nonregsrnet} & TGRS'22 & 33.47 & 0.949 & 0.131 \\
        	MoE-PNP~\cite{qu2024principle} & TNNLS'24 & 33.97 & 0.957 & 0.074\\
        HSIFN~\cite{lai2024hyperspectral} & TNNLS'24 & 41.14 & 0.983 & 0.041 \\
        LRTN~\cite{liu2025low} & IJCV'25 & 37.58 & 0.975 & 0.061\\
        SRLF~\cite{liu2025selective} & CVPR'25 & 38.75 & 0.977 & 0.041 \\
        SSCH-S~\cite{zhang2025unaligned} & IJCV'25 & \underline{41.38} & \textbf{0.987} & 0.031 \\ \rowcolor{graycolor}
        Ours & - & \textbf{41.84} & \underline{0.986} & \textbf{0.025} \\
        \bottomrule[1.2pt]
    \end{tabular}
\end{table}

\begin{table*}[t]
    \centering
    \small
    \caption{Quantitative comparison of various methods under several scale factors on the real dataset, REAL~\cite{lai2024hyperspectral}. PSNR is in dB. The best and second best results are highlighted in \textbf{bold} and \underline{underline}, respectively.}
    \label{tab:result_on_real}
    \renewcommand\arraystretch{0.9} 
    \setlength{\tabcolsep}{4pt} 
    \small
    \begin{tabular}{l|c|ccc|ccc|ccc|cc}
        \toprule[1.2pt]
		\multirow{2}{*}{\textbf{Method}} & \multirow{2}{*}{\textbf{Venue}} & \multicolumn{3}{c|}{\textbf{Scale Factor $\times4$}} & \multicolumn{3}{c|}{\textbf{Scale Factor $\times8$}} & \multicolumn{3}{c|}{\textbf{Scale Factor $\times16$}} & \textbf{Params} & \textbf{FLOPs}\\
        &  & PSNR$\uparrow$ & SSIM$\uparrow$ & SAM$\downarrow$ & PSNR$\uparrow$ & SSIM$\uparrow$ & SAM$\downarrow$ & PSNR$\uparrow$ & SSIM$\uparrow$ & SAM$\downarrow$ & \textbf{(M)} & \textbf{(G)} \\
        \midrule
        Bicubic & - & 34.07 & 0.941 & 0.042 & 28.43 & 0.870 & 0.058 & 24.84 & 0.815 & 0.086 & - & - \\
        SSPSR~\cite{jiang2020learning} & TCI'20 & 39.04 & 0.976 & 0.040 & 31.13 & 0.906 & 0.069 & 27.20 & 0.851 & \underline{0.077} & 20.82 & 180.40 \\
        MCNet~\cite{li2020mixed} & RS'20 & 39.07 & 0.974 & 0.038 & 31.31 & 0.904 & 0.063 & 27.04 & 0.845 & 0.088 & 2.20 & 280.30 \\
        BiQRNN~\cite{fu2021bidirectional} & JSTAR'21 & 37.80 & 0.969 & 0.044 & 31.66 & 0.908 & 0.060 & 26.53 & 0.832 & 0.111 & 1.30 & 468.70 \\
        ESSA~\cite{zhang2023essaformer} & ICCV'23 & 39.94 & 0.982 & 0.040 & 31.85 & 0.913 & 0.064 & 26.70 & 0.918 & 0.111 & 11.10 & 78.14 \\
        Optimized~\cite{fu2019hyperspectral} & CVPR'19 & 27.26 & 0.916 & 0.189 & 26.99 & 0.909 & 0.195 & 26.93 & 0.912 & 0.220  & 0.20 & 14.17 \\
        NonReg~\cite{zheng2022nonregsrnet} & TGRS'22 & 31.60 & 0.951 & 0.081 & 30.00 & 0.923 & 0.167 & 29.09 & 0.904 & 0.584 & 2.00 & 161.61 \\
        	MoE-PNP~\cite{qu2024principle} & TNNLS'24 & 32.57 & 0.951 & 0.101 & 31.12 & 0.946 & 0.119 & 29.91 & 0.934 & 0.121 & 44.67 & 3040.81 \\
        HSIFN~\cite{lai2024hyperspectral} & TNNLS'24 & 40.15 & 0.982 & 0.037 & 34.39 & 0.953 & \underline{0.052} & 30.07 & 0.939 & 0.100 & 21.01 & 594.10 \\
        LRTN~\cite{liu2025low} & IJCV'25 & 38.62 & 0.978 & 0.054 & 34.79 & 0.957 & 0.066 & 31.41 & 0.937 & 0.081 & 3.79 & 33.15 \\
        SRLF~\cite{liu2025selective} & CVPR'25 & 39.06 & 0.978 & 0.043 & 34.63 & 0.953 & 0.064 & 31.12 & 0.929 & 0.091 & 1.33 & 20.31 \\
        SSCH-S~\cite{zhang2025unaligned} & IJCV'25 & \underline{41.16} & \underline{0.986} & \underline{0.036} & \underline{36.19} & \underline{0.969} & \underline{0.052} & \underline{31.91} & \underline{0.940} & 0.084 & 11.01 & 165.68 \\ \rowcolor{graycolor}
        Ours & - & \textbf{42.05} & \textbf{0.988} & \textbf{0.033} & \textbf{37.23} & \textbf{0.972} & \textbf{0.046} & \textbf{32.28} & \textbf{0.942} & \textbf{0.065} & 5.94 & 96.17 \\
        \bottomrule[1.2pt]
    \end{tabular}
\end{table*}

\begin{figure*}
 \centering
\small
\hspace{1mm}
 \setlength{\tabcolsep}{0.1cm}
\begin{minipage}[l]{0.22\linewidth}
\begin{flushleft}
 \vspace{-0.1mm}
  \hspace{-3mm}
\includegraphics[width=1.045\linewidth]{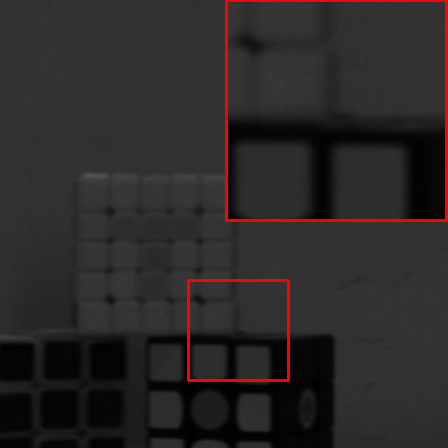}\\
   \begin{tabular}[c]{@{}c@{}}\hspace{2mm}Ground Truth (REAL~\cite{lai2024hyperspectral})\end{tabular}
  \end{flushleft}
 \end{minipage}
 \hspace{-1mm}
 \begin{minipage}[t]{0.5\linewidth}
  \begin{tabular}{cccccc}
   \includegraphics[width=0.2\linewidth]{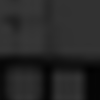}
  & \includegraphics[width=0.2\linewidth]{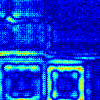}
  & \includegraphics[width=0.2\linewidth]{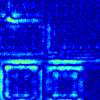}
  & \includegraphics[width=0.2\linewidth]{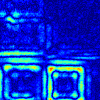}
  & \includegraphics[width=0.2\linewidth]{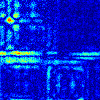}
  & \includegraphics[width=0.2\linewidth]{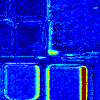}
   \\
	Input ($\times 8$) & SSPSR~\cite{jiang2020learning}  & MCNet~\cite{li2020mixed} & BiQRNN~\cite{fu2021bidirectional} & ESSA~\cite{zhang2023essaformer}  & NonReg~\cite{zheng2022nonregsrnet} \\

 \includegraphics[width=0.2\linewidth]{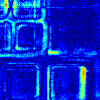}
  &\includegraphics[width=0.2\linewidth]{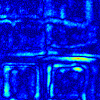}
  &\includegraphics[width=0.2\linewidth]{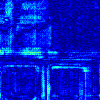}
  &\includegraphics[width=0.2\linewidth]{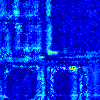}
  &\includegraphics[width=0.2\linewidth]{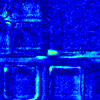}
  &\includegraphics[width=0.2\linewidth]{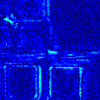}
  
  \\      
	MoE-PNP~\cite{qu2024principle} & HSIFN~\cite{lai2024hyperspectral} & LRTN~\cite{liu2025low}  & SRLF~\cite{liu2025selective} & SSCH-S~\cite{zhang2025unaligned} & Ours
  \end{tabular}
 \end{minipage}
 \hspace{22mm}
 \begin{minipage}[l]{0.1\linewidth}
 \begin{flushright}
    \vspace{-2mm}
     \includegraphics[width=0.54\linewidth]{figures/color_bar_0.1.png}
 \end{flushright}
 \end{minipage}


\hspace{2mm}
 \begin{minipage}[l]{0.22\linewidth}
\begin{flushleft}
 \vspace{-0.1mm}
  \hspace{-3mm}
\includegraphics[width=1.045\linewidth]{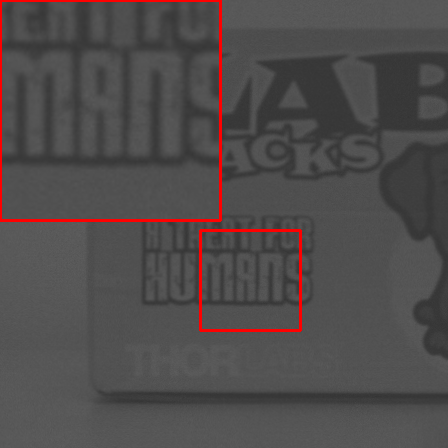}\\
   \begin{tabular}[c]{@{}c@{}}\hspace{2mm}Ground Truth (REAL~\cite{lai2024hyperspectral})\end{tabular}
  \end{flushleft}
 \end{minipage}
 \hspace{-1mm}
 \begin{minipage}[t]{0.5\linewidth}
  \begin{tabular}{cccccc}
   \includegraphics[width=0.2\linewidth]{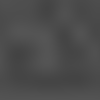}
  & \includegraphics[width=0.2\linewidth]{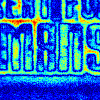}
  & \includegraphics[width=0.2\linewidth]{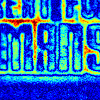}
  & \includegraphics[width=0.2\linewidth]{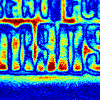}
  & \includegraphics[width=0.2\linewidth]{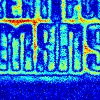}
  & \includegraphics[width=0.2\linewidth]{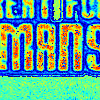}
   \\
	Input ($\times 16$) & SSPSR~\cite{jiang2020learning}  & MCNet~\cite{li2020mixed} & BiQRNN~\cite{fu2021bidirectional} & ESSA~\cite{zhang2023essaformer}  & NonReg~\cite{zheng2022nonregsrnet} \\
   
 \includegraphics[width=0.2\linewidth]{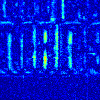}
  &\includegraphics[width=0.2\linewidth]{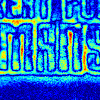}
  &\includegraphics[width=0.2\linewidth]{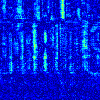}
  &\includegraphics[width=0.2\linewidth]{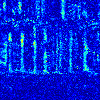}
  &\includegraphics[width=0.2\linewidth]{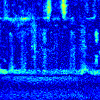}
  &\includegraphics[width=0.2\linewidth]{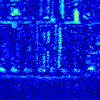}
  
  \\      
	MoE-PNP~\cite{qu2024principle} & HSIFN~\cite{lai2024hyperspectral} & LRTN~\cite{liu2025low}  & SRLF~\cite{liu2025selective} & SSCH-S~\cite{zhang2025unaligned} & Ours  \end{tabular}
 \end{minipage}
 \hspace{22mm}
 \begin{minipage}[l]{0.1\linewidth}
 \begin{flushright}
    \vspace{-2mm}
     \includegraphics[width=0.54\linewidth]{figures/color_bar_0.1.png}
 \end{flushright}
 \end{minipage}

 \caption{Visual comparison on the real dataset, REAL~\cite{lai2024hyperspectral}. The ground truth and input HSIs are shown with band 20. The results of different methods under the scale factor ($\times4$, $\times8$) on the REAL dataset with the error maps.}
 \label{fig:real visual results} 
\end{figure*}

\noindent\textbf{{Results on Simulated Dataset}}.
Tab.~\ref{tab:result_on_icvl} shows a quantitative comparison on the simulated ICVL dataset under a scale factor of $\times4$. As shown, our model achieves a leading PSNR of 41.84 dB and a leading SAM of 0.025, demonstrating its superior capability in recovering both spatial details and spectral signatures. This performance surpasses recent methods, including SSCH-S~\cite{zhang2025unaligned} and HSIFN~\cite{lai2024hyperspectral}, confirming the advantages of our proposed approach.

As shown in Fig.~\ref{fig:icvl visual results}, the error maps represent the absolute pixel-wise difference between each output and the ground truth. It is visually evident that our proposed method achieves a substantially lower error compared to all competing approaches. Methods such as NonReg~\cite{zheng2022nonregsrnet} and MoE-PNP~\cite{qu2024principle} suffer from significant errors, while even recent SOTA methods like HSIFN~\cite{lai2024hyperspectral} and SSCH-S~\cite{zhang2025unaligned} exhibit noticeable high-error patterns along sharp edges. In contrast, the error map of our method is almost dark blue, demonstrating its capability in accurately recovering both overall structure and high-frequency details. More details about other factors are shown in supplementary material.

\noindent\textbf{{Results on Real Dataset}}.
Tab.~\ref{tab:result_on_real} shows a comprehensive quantitative comparison on the real-world dataset, REAL, across scale factors of $\times4$, $\times8$, and $\times16$. Our method consistently outperforms all previous methods across all scales and metrics. Notably, at the most challenging $\times16$ scale, our model achieves a PSNR of 32.28 dB, surpassing the second-best method, SSCH-S~\cite{zhang2025unaligned}, by 0.37 dB. This trend of superior performance is maintained at the $\times$8 and $\times$4 scales, where we achieve PSNR gains of 1.04 dB and 0.89 dB over the second-best, respectively, demonstrating the robustness and effectiveness of our approach in real-world scenarios. Furthermore, our model achieves this leading performance with higher efficiency. It utilizes only 5.94M parameters and 96.17G FLOPs, which is approximately half the parameters and 42\% fewer FLOPs than SSCH-S. This highlights our excellent balance between super-resolution accuracy and computational efficiency.

As illustrated in Fig.~\ref{fig:real visual results}, our method consistently achieves a better visual reconstruction quality across multiple scale factors ($\times$4 and $\times$8). This indicates a more accurate recovery of challenging high-frequency details, such as architectural lines and sharp object edges, where other methods tend to leave residual artifacts, further validating the practical effectiveness of our proposed framework. 
More visualizations are included in supplementary material.

\begin{table}[t]
   \setlength{\tabcolsep}{0.05cm}
   \renewcommand\arraystretch{1}
      \centering
      \small
      \caption{Ablation study of the proposed components. Results on the REAL dataset at scale factor $\times4$.}
   	\label{tab:break-down ablation}
      \begin{tabular}{cccc|cccc}
      \toprule[1.2pt]
      Unmix & SCACA & CFDA & SCMF & PSNR$\uparrow$ & SSIM$\uparrow$ & SAM$\downarrow$ & Params \\ \midrule
      \ding{55} & \ding{55} & \ding{55} & \ding{55} & 41.26 & 0.984 & 0.036 & 5.08M \\
      \ding{51} & \ding{55} & \ding{55} & \ding{55} & 41.41 & 0.986 & 0.034 & 5.05M \\ 
      \ding{51} & \ding{51} & \ding{55} & \ding{55} & 41.66 & 0.987 & 0.034 & 4.85M \\ 
      \ding{51} & \ding{51} & \ding{51} & \ding{55} & 41.95 & 0.988 & \textbf{0.033} & 5.85M \\ \rowcolor{graycolor}
      \ding{51} & \ding{51} & \ding{51} & \ding{51} & \textbf{42.05} & \textbf{0.988} & \textbf{0.033} & 5.94M \\
      \bottomrule[1.2pt]
      \end{tabular}
\end{table} 

\begin{table}[t]
   \setlength{\tabcolsep}{0.24cm}
       \centering
       \small
       \caption{Ablation study on the CFDA module. Results on the REAL dataset at scale factor $\times4$.}
       \label{tab:ablation-CFDA}
       \begin{tabular}{c|cccc}
        \toprule[1.2pt]
        \textbf{Method} & PSNR$\uparrow$ & SSIM$\uparrow$ & SAM$\downarrow$ & Params \\ \midrule
        w/o Aggregation & 41.66 & 0.987 & 0.034 & 4.85M \\
        w/ DCNv2~\cite{zhu2019deformable} & 41.80 & 0.987 & \textbf{0.033} & 5.78M \\ \rowcolor{graycolor}
        w/ CFDA & \textbf{41.95} & \textbf{0.988} & \textbf{0.033} & 5.85M \\
        \bottomrule[1.2pt]
        \end{tabular}
\end{table}

\subsection{Ablation Study}

We conduct a series of ablation studies to validate the effectiveness of our proposed components. 
All ablation experiments are performed on the REAL dataset at a $\times4$ scale factor unless otherwise specified. 
Hyperparameter analysis is provided in supplementary material.

As shown in Tab.~\ref{tab:break-down ablation}, we incrementally add our core modules to a baseline with the self-attention architecture. The introduction of the \textbf{Unmix} strategy provides a strong foundation (41.41 dB of PSNR and 0.036 of SAM). Subsequently, integrating the \textbf{SCACA} module for feature refinement and the \textbf{CFDA} for aggregation boosts the PSNR to 41.66 dB and 41.95 dB, respectively, underscoring the importance of robust feature aggregation and refinement. The final addition of the \textbf{SCMF} module further improves performance to 42.05 dB. The improvements validate that each component contributes effectively to the final HSI.


\noindent\textbf{{Effect of the CFDA Module}}.
Tab.~\ref{tab:ablation-CFDA} specifically analyzes the contribution of our proposed aggregation module. Compared to a baseline without aggregation (41.66 dB), a generic DCNv2~\cite{zhu2019deformable} improves the PSNR to 41.85 dB. Our proposed CFDA module further elevates the performance to 41.95 dB. 
Besides, as shown in Fig.~\ref{fig:Deformable}, we provide a visual comparison of fused feature maps generated by different aggregation modules with the same attention block.
The features aggregated using DCNv2 suffer from severe artifacts and blurred text, whereas our proposed module produces a much cleaner representation with sharper details.
Both the quantitative metrics and the visual comparisons underscore the efficacy of our CFDA module.


\begin{figure}
\begin{center}
\includegraphics[width=0.9\linewidth]{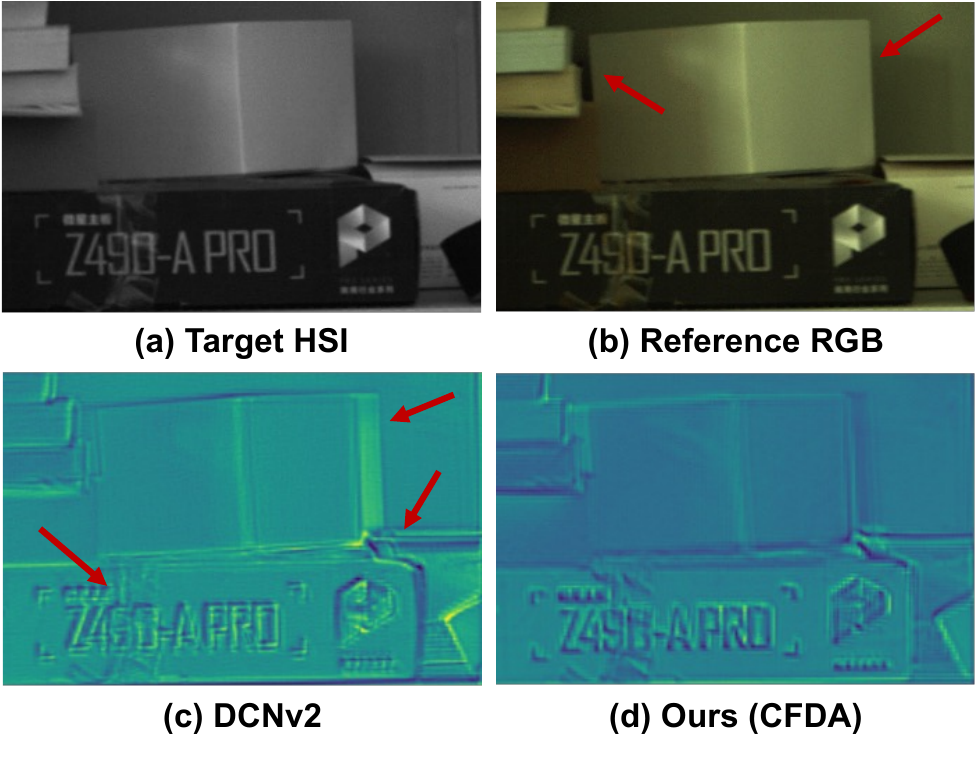}
\end{center}
	\vspace{-7mm}
   \caption{Visual comparison of fused feature maps generated by different aggregation modules, followed by an identical attention block. Note the severe artifacts and blurred text produced by the standard DCNv2~\cite{zhu2019deformable}, in contrast to the superior clarity of the features from our CFDA.}
\label{fig:Deformable}
\end{figure}

\noindent\textbf{{Effect of the SCMF Module}}.
To further validate the effectiveness of the SCMF module, we perform comparative experiments across various scale factors, with the results shown in Tab.~\ref{tab:ablation-SCMF}.
The inclusion of SCMF yields consistent performance gains across all scale factors. Notably, the improvement becomes more pronounced as the super-resolution task becomes more challenging, with a PSNR gain of 0.38 dB at the $\times16$ scale. This highlights the crucial role of SCMF in adaptively fusing multi-scale features for high-fidelity reconstruction.

\begin{table}[t]
   \setlength{\tabcolsep}{0.04cm}
   \renewcommand\arraystretch{1}
       \centering
       \small
       \caption{Ablation study on the SCMF module.}
       \label{tab:ablation-SCMF}
        	\begin{tabular}{c|cc|cc|cc|c}
        \toprule[1.2pt]
        \multirow{2}{*}{\textbf{Method}} & \multicolumn{2}{c|}{\textbf{$\times4$}} & \multicolumn{2}{c|}{\textbf{$\times8$}} & \multicolumn{2}{c|}{\textbf{$\times16$}} & \multirow{2}{*}{Params} \\
        & PSNR$\uparrow$ & SAM$\downarrow$ & PSNR$\uparrow$ & SAM$\downarrow$ & PSNR$\uparrow$ & SAM$\downarrow$ \\ \midrule
        w/o SCMF & 41.95 & 0.033 & 37.02 & 0.047 & 31.90 & 0.068 & 5.85M \\ \rowcolor{graycolor}
        w/ SCMF & \textbf{42.05} & \textbf{0.033} & \textbf{37.23} & \textbf{0.046} & \textbf{32.28} & \textbf{0.065} & 5.94M \\
        \bottomrule[1.2pt]
        \end{tabular}
\end{table}

\section{Conclusion}

In this work, we propose an unmixing-based fusion framework that addresses the critical challenges of explicit alignment (distortions or artifacts) and spatial-spectral coupling in unregistered HSI super-resolution. By leveraging a spectral unmixing paradigm, our approach decouples the problem, enabling robust aggregation via our Coarse-to-Fine Deformable Aggregation (CFDA) module and effective feature refinement with our Spatial-Channel Abundance Cross-Attention (SCACA) block. Finally, our Spatial-Channel Modulated Fusion (SCMF) module ensures the high-fidelity reconstruction of both spatial details and spectral signatures in the final super-resolved HSI. The experimental results clearly demonstrate that our method outperforms other state-of-the-art methods in super-resolution accuracy and visual quality. This highlights the significant potential of our unmixing-based fusion framework to advance the practical applicability of reference-based HSI super-resolution.

\section*{Acknowledgments}
This work was supported by the National Natural Science Foundation of China (62331006 and 62506108), and the Fundamental Research Funds for the Central Universities.


{
    \small
    \bibliographystyle{ieeenat_fullname}
    \bibliography{main}
}


\end{document}